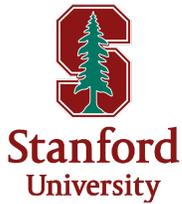
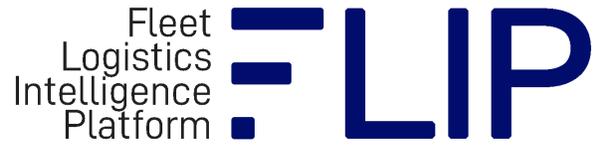

# Predicting Breakdown Risk Based on Historical Maintenance Data for Air Force Ground Vehicles

**Jeff Jang, Dilan Nana, Jack Hochschild, Jordi Vila Hernandez de Lorenzo**

Advisor: Stanford Professor Stefan Wager

**Abstract:**
Unscheduled maintenance has contributed to longer downtime for vehicles and increased costs for Logistic Readiness Squadrons (LRSs) in the Air Force. When vehicles are in need of repair outside of their scheduled time, depending on their priority level, the entire squadron's slated repair schedule is transformed negatively. The repercussions of unscheduled maintenance are specifically seen in the increase of man hours required to maintain vehicles that should have been working well: this can include more man hours spent on maintenance itself, waiting for parts to arrive, hours spent re-organizing the repair schedule, and more. The dominant trend in the current maintenance system at LRSs is that they do not have predictive maintenance infrastructure to counteract the influx of unscheduled repairs they experience currently, and as a result, their readiness and performance levels are lower than desired.

We use data pulled from the Defense Property and Accountability System (DPAS), that the LRSs currently use to store their vehicle maintenance information. Using historical vehicle maintenance data we receive from DPAS, we apply three different algorithms independently to construct an accurate predictive system to optimize maintenance schedules at any given time. Through the application of Logistics Regression, Random Forest, and Gradient Boosted Trees algorithms, we found that a Logistic Regression algorithm, fitted to our data, produced the most accurate results. Our findings indicate that not only would continuing the use of Logistic Regression be prudent for our research purposes, but that there is opportunity to further tune and optimize our Logistic Regression model for higher accuracy.

# 1. Introduction
## 1.1. Background

At each of the United States Air Force's 100+ bases around the world, large fleets of ground vehicles ensure continuity of the mission. At each base, hundreds of vehicles are used for everything from security forces and transporting personnel to towing and refueling aircraft. Certain bases operate upwards of 1000 vehicles. At every base, the local Logistics Readiness Squadron (LRS) owns all vehicles and is tasked with maintaining them for other users on base.

In contrast to aircraft maintenance, vehicle maintenance is notoriously underfunded in the Air Force. As a result, fleet management practices fall short of many commercial vehicle fleet operators. The 4th Logistics Readiness Squadron (LRS) at Seymour Johnson Air Force Base (AFB) enlisted the assistance of the authors to modernize their fleet management via Stanford's Hacking 4 Defense (H4D) course. We have been engaged with them in this effort since March 2021.

The Mission Essential Level (MEL) prescribes the minimum number of vehicles (by type) necessary for the base to support its flight operations. Therefore, the MEL serves as a critical metric for the LRS to prioritize their maintenance operations. For example, the 4th Logistic Readiness Squadron at Seymour Johnson AFB has six hydrant fueling trucks, and the MEL is three fueling trucks. When several fueling trucks are in maintenance, it is important to monitor maintenance schedules and prioritize resources to ensure that the number of operational vehicles never drops below MEL.

Commanders of a base's units annually agree to a MEL for each vehicle type they operate: for example, in 2021 Seymour Johnson's 916th Air Refueling Wing agreed that of its 2 assigned deicers, the Mission Essential Level to enable the continuation of the mission is considered to be 1 operational deicer. Therefore, if both deicers break down simultaneously, the LRS must prioritize resources to prevent impacting the mission.

Our team created a software tool called FLIP: Fleet Logistics Intelligence Platform, and deployed a beta version to the 4th LRS. The initial benefit of the tool is automated creation of business intelligence dashboards, which were previously being put together manually using cumbersome Excel spreadsheets.

## 1.2. Objective

Our objective is to improve the Air Force Fleet manager's ability to predict and prepare for future maintenance work orders in order to create a streamlined workflow. The LRS uses a platform called Defense Property Accountability System (DPAS) to manage and maintain hundreds of ground vehicles at each Air Force Base. DPAS acts as the Air Force's data repository for vehicle work orders.

The LRS is mandated to use DPAS to create, update, and close all of their repair work orders. There are 69 fields of data for each vehicle work order and one work order may also have several sub-work orders associated with it. However, DPAS does not include basic necessary tools for fleet management, including data analytics, macroscopic reporting, and scheduling.

Each of the LRS fleet management personnel we spoke with voiced their frustration with the stop-gap solutions they were forced to create and use on a daily basis to manage their vehicle fleets. They consistently emphasized the following flaws:

1. **Time-consuming**: While some of these spreadsheets use macros built in Microsoft Access or Excel to automate, much of the work needs to be done manually by the LRS's Fleet Management & Analysis (FM&A) section. A survey we conducted found that FM&A personnel spend 30-45 hours per week manually processing this data.
2. **Prone to human error**: Because a human is manually loading and processing much of this data, errors can easily propagate across the tools each airmen builds for his/her LRS.
3. **Lack of standardization and shared knowledge**: Each of the 100+ US Air Force bases has developed siloed tools to report on health and status of their fleet. There is no knowledge flow across bases, FM&A and Commanders of each LRS are not benefiting from tools to track metrics developed by airmen at other bases.
4. **Outdated data**: Because it is time consuming to make the data digestible, there is a limit on how frequently these spreadsheets can be refreshed or new data insights can be added. Additionally, because of the delays associated with moving up the chain of command, when a commander sees metrics generated from these manual analyses, they are likely several days old, which impacts planning and decreases mission readiness levels.
5. **Reliant on "Excel expert"**: Constructing and maintaining these complex spreadsheets requires a local "Excel expert" and the consequences can be disastrous when he/she is relocated to another base. Additionally, inadvertent cell edits can break the functionality of the whole spreadsheet and generate erroneous insights about the status of the fleet.
6. **Reactive, not proactive**: All of these work-around solutions track existing work orders and offer no functionality for forecasting upcoming scheduled maintenance based on past performance.

Our ultimate goal is to create a business intelligence dashboard that takes in current and past vehicle repair data from DPAS and automates otherwise manual analyses in a unified and standardized manner across all Air Force bases. The next step of our product development is creating predictive maintenance algorithms to allow our Air Force customers to be able to predict when a part will fail and ensure that the parts needed to fix the failure are in stock. Having this data in hand, we want to estimate when each breakdown is expected to happen, and notify the airmen to replace the part before it breaks.

## 2. Methods
### 2.1. Dataset

DPAS: Source of the dataset

The datasets that we have been working with were obtained from the centralized data repository DPAS, which for our purposes, acts as a database where airmen across the Air Force regularly update vehicle information and repair status.

DPAS' first version was released in 2011, but it wasn't until a few years later that it gained major adoption across all Air Force bases. To date, all of the 100+ US Air Force bases are using DPAS on a daily basis. Airmen in the field manually upload the latest information of all the vehicle assets in their possession; including the repairs performed on each vehicle and the hours it took to complete them.

Dataset chosen: Sub-work order inquiry

DPAS offers the capability to pull various tables from the database applying several filters to minimize the file size. The dataset that we have used for all of our models is the "sub-work order inquiry", where each row is a sub-work order entry, which contains 69 unique fields (columns), the most relevant ones are discussed below. The data inquiry request to DPAS is done through a web application, only accessible from a Government computer connected to the AF Network. The file format type to download can be selected between a comma separated value (CSV) or Excel file. Our algorithms can accept both types of data inquiry pulls, CSV and Excel.

To reduce human error and the time it takes US Air Force personnel to select and download the raw data file from DPAS (no more than 2 minutes), we have programmed our software such that the dataset that we take as an input is a full inquiry of all the information on the database, with all 69 data fields selected (although we don't use all of them to run our predictive analysis).

While the dataset format is standard across all US Air Force bases, the data on each dataset is unique to each base as it is vehicle-specific. Each dataset is unique to each base, and contains historical sub-work orders for all the vehicles that have been repaired on that base since the DPAS system was adopted.

To date, we have obtained datasets from more than 10 US Air Force bases, and some of them contain data with over 50,000 rows (or sub-work order jobs) performed over the last 5 years. This widespread adoption and amounts of data plays in our favor as the datasets available to us contain a million of data points and historical data that we use to better train our models and verify the accuracy of our predictions.

Out of the 69 unique fields (or columns) from the "sub work order" dataset inquiry we use the ones that are more relevant to compute our predictions, here are some of the most important ones and what they mean:

| Field Name | Definition |
|---|---|
| Work Order ID | Unique identifier for the Word order ID. Each time a vehicle gets into the shop, a "master" Work Order entry is created and depending on how many subtasks have to be performed on the repair, new "sub work orders IDs" will be added for the same work order ID. |
| Sub Work Order Id | Each Work Order ID can have multiple Sub work orders, or independent tasks, that have to be performed on the vehicle (ie; oil change, change brake pads, etc.). Each of these tasks would get a different "sub work order ID" assigned (01, 02, etc) but all would share the same parent Work Order ID. |
| Approval Dt | Date that the vehicle was checked into the shop. Date that the repair starts. |
| Asset Id | Unique vehicle identifier across the Air Force (ie; AF08l00508). This field also indicates the year in which the vehicle was acquired, for instance AF08..means that vehicle was acquired in 2008. Since generally vehicles are acquired new, the asset ID can be used to extract the age of the vehicle. |
| Closed Dt | Date when the sub work order was closed (repair completed) |
| Item Desc | Vehicle general description (Truck lift fork, Semitrailer low-bed, Truck cargo, Truck Tractor, etc.) |
| Asset LIN/TAMCN | Also called Management Code (MGMT CD) is one of the 550 Vehicle types (Crossover, Hybrid Sedan, 4x2 Cargo Van, etc) |
| Equipment Pool | Unit who owns the vehicle inside the base (31 AMXS/555 AMU, LRS VEHICLE OPS, LFS FUELS, etc) |
| Maint Team Name | Mechanic shop that performed the repair inside the base |
| Estbd Dt/Time | When the repair was registered in the database |
| Work Plan Type CD | 27 possible work plans (Adjustment, Alignment, Calibration, Major |

|  | Repair, Prev, Troubleshoot, etc). Out of the 27 possible work plans, "Prev", is the only one that corresponds to scheduled maintenance (oil changes, brake inspections, etc. based on mileage or usage). The remaining 26 work plans correspond to services that were not planned, or unscheduled maintenance. Our goal is to estimate those unplanned breakdowns using historical data. |
|---|---|

Hypotheses for using these features for the regressions

Our main Hypothesis going into this analysis was that the most accurate predictions of vehicle failures (when subwork orders are opened for a specific Asset ID) would be heavily dependent on these parameters:
1. The age of the vehicle, older vehicles should break more often than newer vehicles.
2. The mileage of the vehicle, as wear and tear increases likelihood of maintenance required.
3. The time elapsed since the last visit to the mechanic shop (when last sub work order was opened for that same Asset ID), the more time since last checkup, the more likely the vehicle is to break down.

A secondary hypothesis was that certain units might require more maintenance in their vehicles than others, that could mean either more sub-work orders opened in the same amount of time or just more total labor hours spent repairing the vehicles owned by that unit.

A caveat to this secondary hypothesis is that even if some units require more maintenance than others, that could be related to the type of vehicle that each unit mostly owns. Ie; Some units might mainly own Heavy weight Loaders, while others might own mainly pickup trucks. Heavy Loaders require more maintenance due to the nature of their usage; loading heavy equipment into planes makes them more susceptible to breakdowns. On the other hand, pickup trucks are used as a transportation means to haul airmen within the base and are less likely to be exposed to strenuous situations.
However, if both units roughly owned the same type of vehicles, this regression could be an indicator of human behavior and how much care each unit takes of their fleet. This could be valuable data to inform accident and abuse habits to LRS' upper management.

Another hypothesis was that using our regression analysis for all vehicles (Asset ID) of the same category (Vehicle type) could flag specific vehicles that have systemic and recurring problems. If that were true, we could use this information to inform decision makers to replace or retire certain assets that are more expensive to maintain than similar vehicles, saving resources to the Air Force.

With these hypotheses in mind, we assembled a historical dataset from Seymour Johnson Air Force Base with over 50,000 row entries of repair data, or over 3 million data points, to validate or invalidate our hypotheses.

To prepare the data for the regressions we construct a time series (in weeks) for each vehicle, with each week populating a "repair flag" which equals 1 if the vehicle was serviced that week, and 0 otherwise.

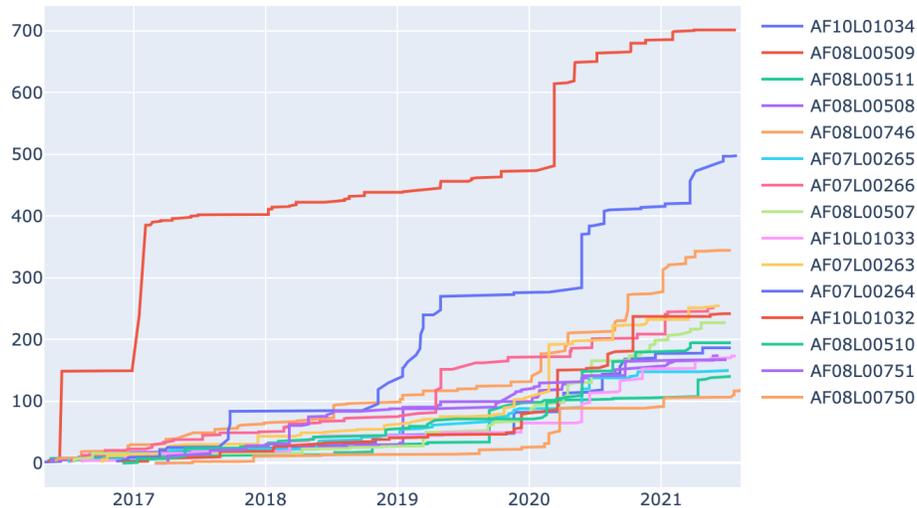

**Figure 1.** Time series of repair hours on a specific vehicle as a function of age/utilization for different vehicles

## 2.2. Regression Models

We tested three different regression algorithms to determine which one would best fit our purpose and the data that we possess. Below are the three different algorithms we tested and our explanations for their respective implementations.

Logistic regression

Logistic Regression seemed to be a perfect algorithm to select and test given our research question: "How can we determine whether a vehicle should be maintained this week or not" and the more specific question "Has this particular vehicle X been repaired this week, yes or no?" Given that we were dealing with a 'yes or no' research question as described above, and we had historical, labelled data, logistic regression seemed fitting.

Our dependent variable was whether the vehicle was repaired in the given week or not or as some logistic regression literature refers to as 'present - 1' or 'absent - 0.' We did not analyze coefficients individually, but going forward, we hypothesize that this would provide clarity or confirmation on the most influential features. From this, we could determine the individual influence of each feature.

To implement the model, we took into account a finite number of independent variables relating to the vehicle, including its ID, its type, the unit to which it was assigned, etc. However, we kept in mind that the more variables we added, the more likely we were to cause the model to overfit, so we looked at our data through the lens of solely what we thought would factor into the question: "Has a particular vehicle been repaired this week, yes or no?"

Given this question, selecting the appropriate features from our data was fairly straightforward: we began by calculating the 'influence' of the features. To do so, we scaled each feature we thought would be influential: vehicle ID, vehicle type, unit to which the vehicle was assigned, weeks since last visit, age of the vehicle, and utilization, to have the same standard deviation. To determine the standard deviation, we looked at each column of the model matrix and divided them by their standard deviation, respectively. After scaling each feature to have the same standard deviation, we then analyzed the magnitude of the coefficients to determine the importance of each feature. [1]

Random Forest
For our second approach to tackling predictive maintenance, we employed the random forest regression algorithm. Due to the 'wisdom of the crowds' innate quality, random forest regression can provide high accuracy, and it has been documented to produce more accurate results than linear models such as logistic regression.

To enable random forest regression, we built decision tree regressors to answer the same questions posed above: "How can we determine whether a vehicle should be maintained this week or not" and "Has this particular vehicle X been repaired this week, yes or no?" Furthermore, to optimize our model, with each node split, we entered the specific number of maximum features.

Gradient Boosted Trees
The third algorithm we tested was the gradient boosted trees algorithm. Our main motivation for selecting this algorithm was that gradient boosting had been seen to improve model performance through interactions wherein it can recursively fit the residual to a weak learner.[2]

This algorithm could result in better performance than random forest if we tuned the parameters well, so we thought we could compare the outcomes between the two algorithms. The problem here is that with gradient boosting, noise can cause overfitting, and our data is quite noisy, with numerous instances of label noise.

---

[1] Schielzeth, H. (2010), Simple means to improve the interpretability of regression coefficients. Methods in Ecology and Evolution, 1: 103-113.
[2] Zhang, Zhongheng et al. "Predictive analytics with gradient boosting in clinical medicine." *Annals of translational medicine* vol. 7,7 (2019): 152. doi:10.21037/atm.2019.03.29

For our gradient boosting, our target outcome variable was if a particular vehicle was "being repaired in a given week or not." For predictors, we used vehicle ID, vehicle type, unit to which the vehicle was assigned, and the other variables enumerated below in the results. We split the data into a 70% training set and a 30% testing set, so that we could fit the model to our 70% training set and then apply the model to the test set. From testing on the test set, we can determine the performance of the model.

The Python Scikit Learn package was used to implement these models.

## 3. Results (& discussion)
### 3.1. Logistic Regression
A logistic regression was performed on the service history dataset. The features used were:
- Vehicle ID
- Vehicle type
- Unit vehicle assigned to
- Weeks since last visit
- Operational weeks (age of vehicle)
- Utilization (miles/hours)

And the outcome is a repair flag of value 0 or 1, which represents whether the vehicle is repaired during that week. The data was split into test/train by taking test data from random points throughout the time series.

The histogram in Figure 2 shows the distribution of the probability that a vehicle will need to be serviced for the test data. The distributions are separated into data points which had a true value of 1 (red) and those with a value of 0 (blue). The distance between the distributions is a measure of predictive ability: the two distributions would have no overlap if the regression were perfectly able to determine whether a vehicle needs servicing on a given day, and conversely equal distributions implies zero predictive ability. For this regression, the mean prediction for false outcomes (no service performed) is 0.062 and for true outcomes (service performed) is 0.093. The ratio of the means is 1.49, meaning the predictions are better than random (ratio would be 1).

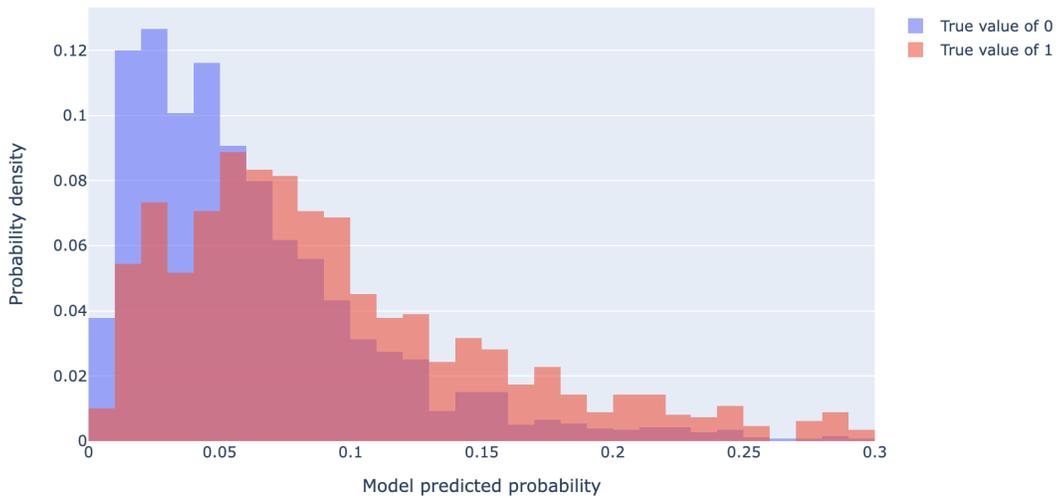

**Figure 2.** Distribution of the probability that a vehicle will need to be serviced for the test data

In an actual use case, this regression would predict breakdowns for some period into the future, rather than at random points interspersed across historical data. Therefore, we instead divided the dataset into test/train using a chronological split: we test on the last 30% of the data. The following histogram shows the same result as Figure 2, but using this alternate method of test/train split. In this case, the mean predictions for false outcomes is 0.070 and for true outcomes is 0.104, giving the same ratio of 1.49.

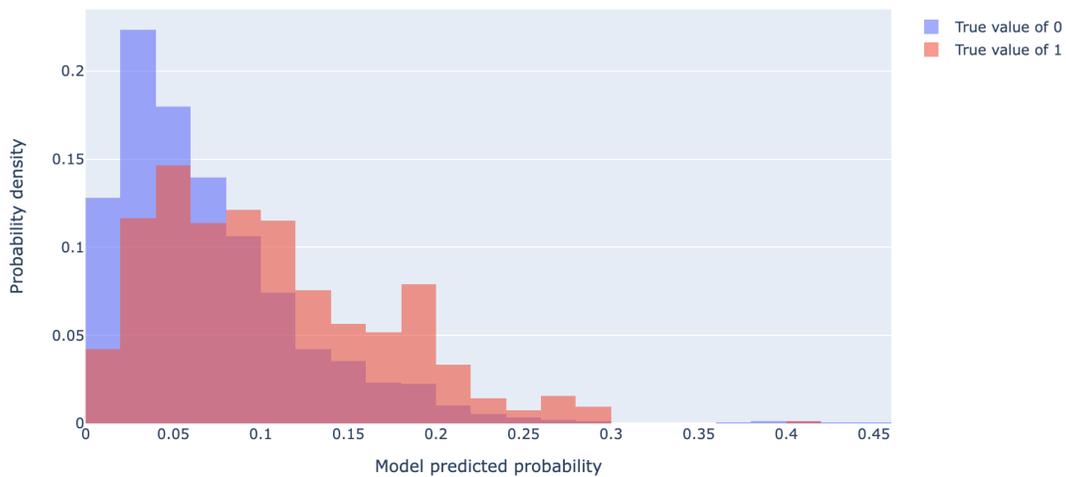

**Figure 3.** Alternate train/test splitting for logistic regression model

We investigated the influence of various features on predictive ability by doing regressions without features. The results are shown in Table 1, which illustrates the ratio of the mean predictions for true/false outcomes is given as a function of features used.

| Vehicle ID | Vehicle Type | Operational weeks | Weeks since last visit | Utilization | Unit Assigned | **Ratio of mean predictions** |
|---|---|---|---|---|---|---|
| ✔ | ✔ | ✔ | ✔ | ✔ | ✔ | **1.49** |
|   | ✔ | ✔ | ✔ | ✔ | ✔ | **1.48** |
|   | ✔ | ✔ | ✔ | ✔ |   | **1.43** |
|   | ✔ | ✔ | ✔ |   |   | **1.41** |
|   | ✔ | ✔ |   |   |   | **1.38** |
|   |   | ✔ |   |   |   | **1.02** |

**Table 1.** Ratio of mean predictions for true/false outcomes given variables selected

As we may have surmised from the plots of repair hours as a function of vehicle age in Figure 4, the age of the vehicle (operational weeks) alone is not a good predictor of breakdowns. However, using the type of vehicle as well as age gives the single largest increase in predictive power (1.38). This is consistent from the results of Figure 5, since certain vehicle types did show a trend towards an increased maintenance burden for older vehicles.

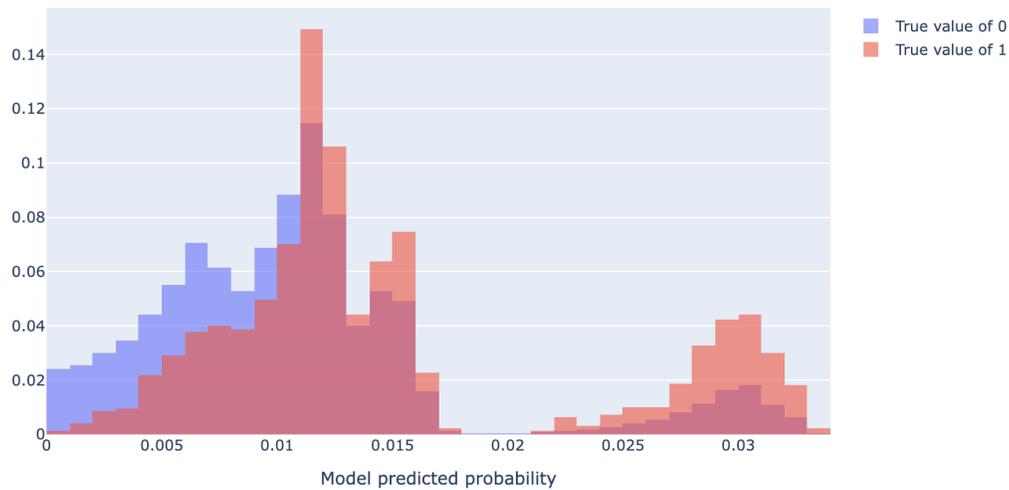

**Figure 4.** Probability density function (PDF) of outcomes using only using repair hours as a function of vehicle age in weeks

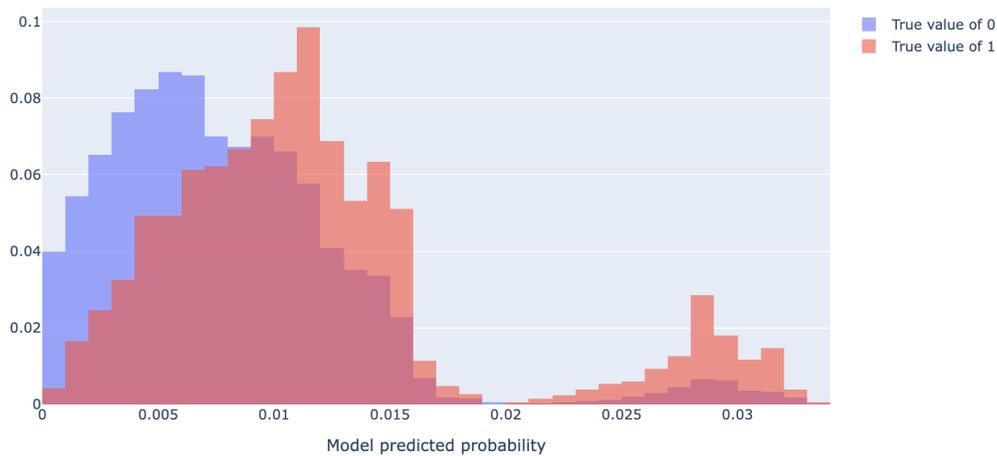

**Figure 5.** PDF of outcome using vehicle age in weeks and vehicle type

An additional notable finding is that the predictive capability does not notably increase when unique Vehicle ID is added as a feature. Practically speaking, this is encouraging since we would be able to make predictions on a new vehicle added to the fleet based on records for vehicles of the same type. Additionally, since Vehicle ID is the most granular of the discrete features, excluding it reduces computational burden for very large datasets.

### 3.2. Other Regressions

We also used Random Forest and Gradient Boosted Trees. For each method, we tuned maximum depth and number of estimators to maximize performance on the test dataset. The histograms of predictions are shown in Figure 6 and Figure 7. Random Forest and Gradient Boosted Trees achieved mean prediction ratios of 1.42 and 1.43, respectively. Therefore these methods underperform compared with Logistic Regression, which achieved a mean ratio of 1.49.

To optimize the Random Forest algorithm, we are contemplating adjusting the number of features. We tried using fewer features and did not see relative improvements to logistic regression. Going forward, if we were to use this algorithm, we would need to figure out how to tune our features optimally, since one could argue that decreasing features would move the model towards higher accuracy, yet we did not experience such in our findings.

Some threats of Random Forest we kept in mind included overfitting as a result of not specifying the maximum depth of trees. To optimize our findings, we tried increasing and decreasing the number of estimators. Since more trees mean a slower learning process but higher accuracy, we are contemplating two options: increasing the learning speed through lowering the number of estimators or increasing accuracy by increasing the number of trees. We ended up selecting 400 for our estimators as a result of tuning our model to optimize output.

To optimize the Gradient Boosted Trees algorithm in the future, we would most likely select a higher learning rate. This could entail selecting a rate in the range of 0.05 to 0.2. Another option we contemplate using in the future would be to lower the learning rate. As we lower the learning rate, simultaneously, we proportionally increase the estimator. Lastly, we hypothesize that through trials, we could figure out, for our given learning rate (40-70 possibly), the optimum number of trees.

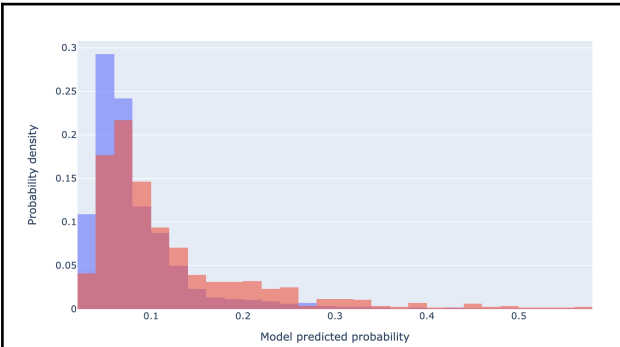 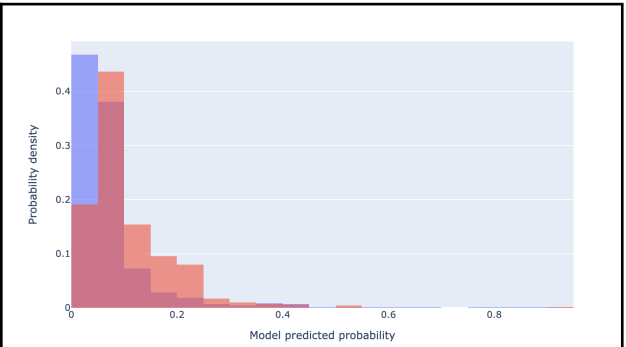

**Figure 6.** Random Forest   **Figure 7.** Gradient Boosted Trees

### 3.3. Evaluation

We tested the model in a way that emulates how it could be used in an Air Force LRS. Each week, we proactively repair the vehicle with the highest probability of breakdown and then return it to service. We repopulate the weeks since the last repair column to reflect the recent service. After doing this for each week during the test dataset, we can evaluate how soon thereafter these proactively repaired vehicles needed to actually be repaired.

The histogram in Figure 8a shows the distributions of 1) when the last time the proactively repaired vehicle was serviced (blue) and 2) how many weeks until the next service (red). The histogram in Figure 8b shows as a baseline the distributions if instead a random vehicle is selected to be repaired each week. Comparing the two, we see that by taking the vehicle with the highest breakdown risk we are more likely to repair it shortly before it breaks down.

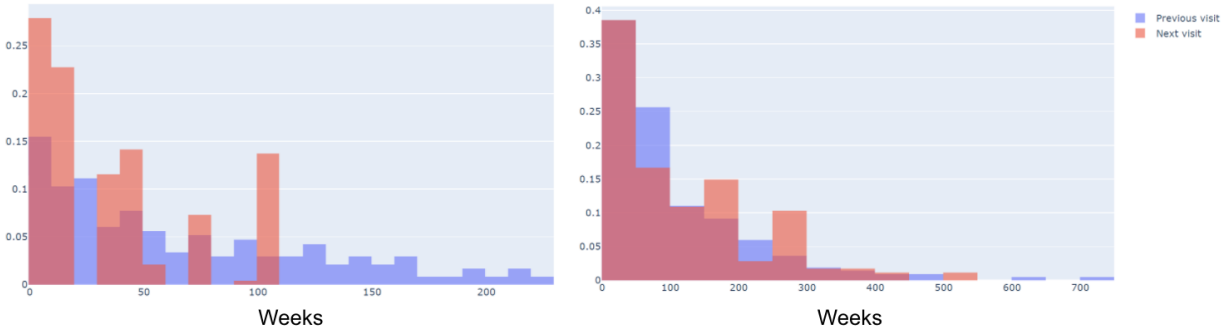

**Figure 8a.** (Left Histogram) Proactively repaired vehicle vs weeks until next service
**Figure 8b.** (Right Histogram) Random vehicle selected to be repaired next week

# 4. Conclusions

During our project, we sought to develop a tool that can serve as a robust business intelligence tool for the Air Force LRS community. This paper specifically aimed at giving FM&A personnel the capability to reasonably determine a vehicle breakdown before it happens. From our preliminary findings, we discovered that the age of the vehicle alone is not sufficient enough to predict vehicle breakdowns. However, by adding the following factors to our logistic regression, we were able to significantly increase the model's predictive capabilities: Vehicle ID, Vehicle Type, Operational weeks, Weeks since last visit (for maintenance), Utilization, and Unit Assigned.

Overall, the analysis of our results show that our predictive models based on logistic regressions allow airmen to predict with 49% higher confidence of a vehicle breakdown than randomly repairing vehicles (reactive maintenance). Building our predictive models using Random Forest and Gradient Boosted Trees did not result in a higher mean prediction ratio than using logistic regressions, but they were also more effective at predictive vehicle breakdowns than randomly repairing vehicles. The Random Forest method resulted in 42% higher confidence and the Gradient Boosted Trees method resulted in 43% higher confidence. In all three methods, by having our model select the vehicle with the highest breakdown risk, based on historical utilization data (normalized by observed utilization), we are more likely to repair it shortly before it breaks down.

For our next steps, first we will investigate quantifying the risk of breakdown by vehicle type and aggregate the failure probabilities by vehicle type as part of our dashboard features. This will allow our Air Force partners to quantify the risk of going below MEL for each vehicle type in the future, which will inform the priorities of their current maintenance operations and increase Mission Readiness.

Secondly, we will work on integrating sensor data into our predictive models to understand whether they can improve the mean prediction ratios of our models that only use analog historical data. The Air Force is trialing telematics for their vehicle fleets, but the sensor data from those trials are not yet available for us to test our models.

Finally, we will continue to develop our predictive models into a functional tool for the Air Force. We envision a scheduling tool, with a 90-day calendar view, that gives suggestions for when vehicles should be called in for maintenance, when the maintenance should be performed, and an estimated time for when the maintenance should be completed, based on our predictive models. It will also be interactive (drag and drop individual work orders to different days), which will give FM&A personnel the flexibility to dynamically change the schedule as needed and observe the impact of those changes. We also plan on further investigating the integration of supply chain processes, so that the parts required to perform maintenance on a vehicle are already in inventory by the time it is called in for maintenance.